\documentclass[a4paper,conference]{IEEEtran}
\usepackage[T1]{fontenc}
\usepackage{graphicx}
\usepackage{multirow}
\usepackage{hyperref}
\usepackage{algorithm}
\usepackage{algorithmic}
\usepackage{amsmath}
\usepackage[backend=biber]{biblatex}

\ifCLASSINFOpdf

\else

\fi

\begin{document}
%
\title{Unsupervised Domain Adaptation for \\ Person Re-Identification through \\ Source-Guided Pseudo-Labeling}

\author{\IEEEauthorblockN{Fabian Dubourvieux$^{\star, \dagger}$, Romaric Audigier$^{\star}$, Angelique Loesch$^{\star}$, Samia Ainouz$^{\dagger}$, Stephane Canu$^{\dagger}$}
\IEEEauthorblockA{$^{\star}$ \textit{CEA, LIST}, \textit{Vision and Learning Lab for Scene Analysis}\\ PC 184, F-91191 Gif-sur-Yvette, France\\
    \{firstname.lastname\}@cea.fr\\}
\IEEEauthorblockA{$^{\dagger}$ \textit{Normandie Univ, INSA Rouen, LITIS}\\ 
Av. de l'Université le Madrillet 76801 Saint Etienne du Rouvray, France\\ \{firstname.lastname\}@insa-rouen.fr\\}
}
%

\maketitle

\begin{abstract}
Person Re-Identification (re-ID) aims at retrieving images of the same person taken by different cameras.  A challenge for re-ID is the performance preservation when a model is used on data of interest (target data) which belong to a different domain from the training data domain (source data). Unsupervised Domain Adaptation (UDA) is an interesting research direction for this challenge as it avoids a costly annotation of the target data. Pseudo-labeling methods achieve the best results in UDA-based re-ID. They incrementally learn with identity pseudo-labels which are initialized by clustering features in the source re-ID encoder space. Surprisingly, labeled source data are discarded after this initialization step. However, we believe that pseudo-labeling could further leverage the labeled source data in order to improve the post-initialization training steps.
In order to improve robustness against erroneous pseudo-labels, we advocate the exploitation of both labeled source data and pseudo-labeled target data during all training iterations. To support our guideline, we introduce a framework which relies on a two-branch architecture optimizing classification and triplet loss based metric learning in source and target domains, respectively, in order to allow \emph{adaptability to the target domain} while ensuring \emph{robustness to noisy pseudo-labels}. Indeed, shared low and mid-level parameters benefit from the source classification and triplet loss signal while high-level parameters of the target branch learn domain-specific features.
Our method is simple enough to be easily combined with existing pseudo-labeling UDA approaches. We show experimentally that it is efficient and improves performance when the base method has no mechanism to deal with pseudo-label noise. And it maintains performance when combined with base method that already manages pseudo-label noise.
Our approach reaches state-of-the-art performance when evaluated on commonly used datasets, Market-1501 and DukeMTMC-reID, and outperforms the state of the art when targeting the bigger and more challenging dataset MSMT.
\end{abstract}


%
\IEEEpeerreviewmaketitle

\section{Introduction}
Person re-identification (re-ID) aims at retrieving images of a person of interest captured by different cameras. It is known as an \emph{open-set problem} because identities (i.e., classes) seen at test time are different from those at training time. Effective representations must be learned in order to discriminate people (i.e., classes) never seen during training. Convolutional Neural Networks (CNNs) have achieved excellent performance on various re-ID datasets, due to their capacity to learn camera invariant and identity discriminative representations robust to appearance changes \cite{luo2019bag}.

In practice, one may want to deploy a re-ID system on a different set of cameras than those used for training data, i.e. on a test set from a different distribution (brightness, colorimetry and angle of the cameras, background...). In this context of \emph{cross-dataset} testing, we observe in practice a sharp drop in re-ID performance due to the \emph{domain gap} \cite{deng2018image}. A solution may be the annotation of larger scale re-ID datasets or the target dataset itself so that a new better performing model can be trained on the target domain. To avoid these costly manual annotations, the research community focused on designing Unsupervised Domain Adaptation (UDA) algorithms. UDA aims at learning an efficient model on the target domain. It only requires labeled training samples from one or multiple domains (source domains) and unlabeled samples from the domain of interest (target domain) on which the model is tested. While extensive research is conducted for classification UDA, the open-set nature of the re-ID task makes it impossible or inefficient to directly apply closed-set approaches designed for classification \cite{panareda2017open}. Therefore, a part of computer vision researchers focuses on the peculiarities of re-ID being an open-set retrieval problem.

In this setting, pseudo-labeling approaches have proven to be the best UDA methods to learn ID-discriminative features for the target domain \cite{ge2020mutual}. These ID pseudo-labels are initially predicted by clustering the target data embedded in an ID-discriminative feature space learned on source data. To do so, pseudo-labeling UDA algorithms alternate between training phases and uour baselinetes of pseudo-labels with the lastly trained model to refine them to a certain extent.
Through this iterative process, the source data is not reused beyond the initialization step of the model. It can be assumed that this choice not to integrate the labeled source data aims at not biasing the learning of ID-discriminative features on target data. One could think that using source data to constrain the training would decrease adaptability to target domain and would degrade the re-ID performance on target domain. We believe that it is possible to subtly exploit the source data and their ground-truth labels to improve the representation on the target domain, while reducing the potential undesirable effects of the domain gap between the source and target data.

Our contribution can be summarized as follows. 
\begin{itemize}
    \item We propose a \emph{source-guided pseudo-labeling} framework to solve \emph{cross-dataset re-ID}. It leverages the labeled source data during pseudo-labeling scheme of all training phases.
    \item The key element of this framework is a two-branch architecture that simultaneously optimizes classification and triplet-based metric learning in both source and target domains, in order to allow \emph{adaptability to the target domain} (high-level parameters of target branch optimized with pseudo-labels) while ensuring \emph{robustness to noisy pseudo-labels} (shared low and mid-level parameters constrained by supervised learning with source labels).
    \item The proposed framework is simple enough to be easily combined with existing pseudo-labeling UDA approaches. Experiments show that it is particularly effective when the base method has no special mechanism to deal with pseudo-label noise. It also improves the stability of re-ID performance with relation to the tricky choice of clustering parameters.
    \item Our framework combined with MMT method outperforms the state-of-the-art on the challenging cross-domain scenario of MSMT dataset.
\end{itemize}

In Section~\ref{related}, we review the related work. Then, we detail our proposed framework in Section~\ref{sec:framework}. The experiment settings are given in Section~\ref{exp} and results are presented in Section~\ref{Results}.

\section{Related Work}
\label{related}

The state of the art contributions for UDA re-ID can be divided into 3 families which leverage the source data differently.

\subsection{Image-to-Image translation}Image-to-Image translation methods \cite{wei2018person} \cite{deng2018image} \cite{qi2019novel} \cite{zhong2018generalizing} are based on learning how to transform images from one domain to another, preserving class information. The majority of these approaches use a CycleGAN \cite{zhu2017unpaired} model to transfer the style of the target images to the source images, while constraining the person's appearance preservation after transfer. The target-style source images are then used to learn in a supervised way a re-ID model for the target domain.
Although these methods fully exploit the source images (in the target style) and their labels, they depend on the quality of image generation and the preservation of identity information after transfer. In practice, the performance obtained by these methods shows that the source images in the target style are not sufficiently representative of the target domain.

\subsection{Domain invariant feature learning} Domain invariant feature learning methods look for a domain invariant discriminative feature space \cite{chang2019disjoint} \cite{lin2018multi}  \cite{li2018adaptation} \cite{li2019cross}. In addition to the supervised ID discriminative loss on source samples, these approaches seek to align the source and target domain feature distributions by penalizing an unsupervised domain discrepancy loss term  \cite{chang2019disjoint} \cite{lin2018multi} or learn domain invariant space by domain feature disentanglement  \cite{li2018adaptation} \cite{li2019cross} sometimes with auxiliary information for supervision (semantic attribute labels, pose labels...) \cite{wang2018transferable}.
As with methods that operate at the pixel level, domain invariant feature learning makes full use of source data and their labels. While they can outperform Image-to-image translation approaches \cite{chang2019disjoint} \cite{chang2019disjoint}, these methods cannot learn efficient target specific features because of domain invariance constraints and the presence of discriminative information (labels) only for the source data.

\subsection{Pseudo-labeling} Pseudo-labeling methods exploit a source-trained model to initialize pseudo-identity labels for target data by clustering their feature representation by this model \cite{yu2019unsupervised} \cite{zhong19enc}. Most of pseudo-labeling UDA methods are based on an iterative paradigm which alternates between optimizing and pseudo-label refinery by uour baselinetes with the lastly optimized model on target images \cite{song2020unsupervised} \cite{zhang2019self} \cite{fu2019self} \cite{tang2019unsupervised}. However, the source biased representation and the clustering introduce some errors that can persist through iterations, be over-fitted by the model and thus degrade the pseudo-label refinery process. To avoid over-fitting the noise of these generated labels, recent work focused on improving Pseudo-labeling methods by improving their robustness to noisy labels using asymmetric co-teaching \cite{yang2019asymmetric} or mutual mean teaching with soft target labels \cite{ge2020mutual}. Even though recent Pseudo-labeling methods achieved near supervised training performance on some re-ID adaptation tasks, they still suffer from sensitivity to clustering or additional hyperparameters which can't be easily estimated in the UDA setting where no labeled target validation set is available. \cite{ge2020mutual} Moreover, we believe that they under-exploit the labeled source dataset since it is discarded after the initialization phase. While UDAP method \cite{song2020unsupervised} uses an additional weight ratio term between source and target to measure feature similarity, it is only used during the pseudo-label predictions by clustering. Moreover, their ablation studies show negligible performance gain and the source samples are still discarded from the optimization iterations.

Our contribution focuses on these pseudo-label iterative methods. In contrast to the existing pseudo-labeling approaches mentioned above, we propose to better leverage the labeled source samples, beyond the initialization phase, in order to further improve their target discriminative features. Contrary to UDAP \cite{song2020unsupervised}, we leverage the source samples as well as their labels directly during the optimization phases of the iterative training.

\section{Proposed framework}
\label{sec:framework}

\begin{figure*}[!t]
\centering
\includegraphics[width=1.5\columnwidth]{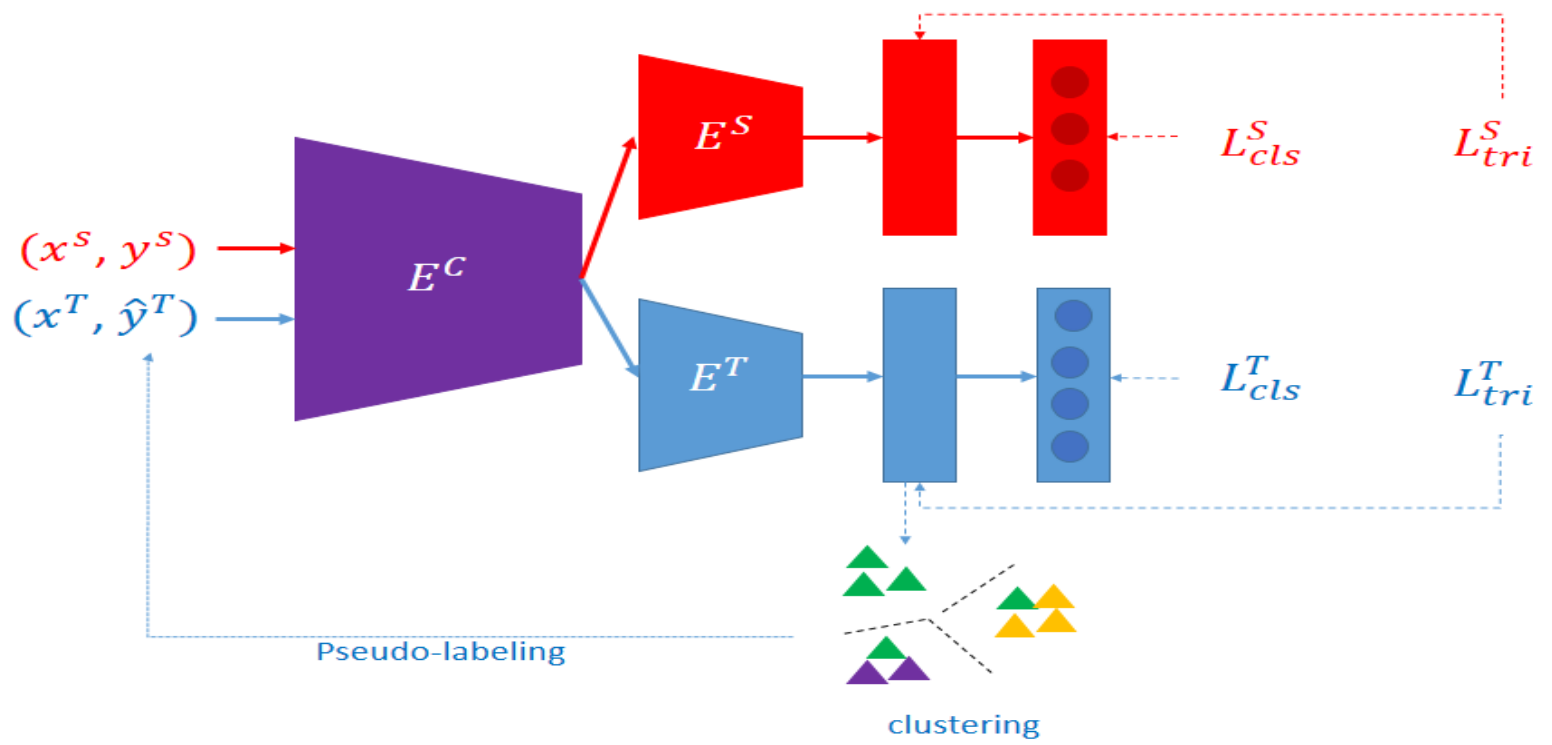}
\caption{Unlike existing pseudo-labeling strategies (1), our framework is a source-guided (3) as it leverages the labeled source dataset during pseudo-label training. It is composed of a shared encoder $E^C$ and two domain-specific branches $E^S$ and $E^T$. The numbers of shared layers and specific layers can vary. Configuration (2) represents the extreme case when all layers are shared and there is no domain-specific branch.}
\label{fig:framework}
\end{figure*}

In this section, we describe our source-guided pseudo label framework. As other pseudo-labeling UDA frameworks, it is composed of two distinct parts:\\
- the supervised training of the initialization model on source: we follow the supervised training guideline from used \cite{luo2019bag} by most pseudo-labeling UDA re-ID \cite{ge2020mutual} \cite{yang2019asymmetric}  \\
- our source guided pseudo-label iterative training to efficiently leverage the source samples

\subsection{Supervised training on source domain for initialization}
Let $D^S = \{ (x^S_k,y^S_k)_{1 \leq k \leq N^S} \}$ be the labeled source training set of $N^S$ samples from $M_{S}$ different people, where $x_{k}$ represents the k-th person image and $y_{k}$ its one-hot identity label. Similarly let $D^T = \{ (x^T_k)_{1 \leq k \leq N^T} \}$ the unlabaled target training set of $N^T$ samples. $E$ denotes a learnable feature encoder.

\subsubsection{Triplet Loss}

To learn re-ID discriminative features, we use the batch hard Tiplet Loss \cite{hermans2017defense} to pull the same ID sample features closer to each other than other ID ones. In a batch of $N$ samples $\{ (x_k,y_k)_{1 \leq k \leq N^S} \}$, it is given by:

\begin{equation}
  \begin{split}  
  L_{tri} = \sum_{i=1}^{N} \max(0, \left \| E(x_{i}) - E(x_{p(i)}) \right \|_{2} \\
- \left \| E(x_{i}) - E(x_{n(i)})  \right \|_{2}
+ m)
\label{triplet}
\end{split}
\end{equation}

where $m$ is a fixed margin, $x_{p(i)}$  and $x_{n(i)}$ are resp. the hardest positive (the farthest batch feature sample to $x_{i}$ for the L2 distance) to  and negative (the closest) samples in the batch for $x_{i}$.\\

\subsubsection{Cross-Entropy Loss}
To further improve the ID discriminativeness of features, we use the classification of ID labels with the Cross-Entropy Loss. For this, we consider a classification layer given by the parameters $W = [ W_1,...,W_{M} ]$ where $M$ is the number of ID in the training set. We use the Softmax classification loss given by:

\begin{equation}
L_{cls} = - \sum_{i=1}^{N}y_i \log(p_i)
\label{cls}
\end{equation}

where $ p_i  = \frac{exp({W_{y_i}E(x_i)})}{\sum_{k=1}^{M}exp(W_{k}E(x_i))}$.
\\

\subsubsection{Initialization phase loss}

By considering a source domain classification layer $W^S = [ W^S_1,...,W^S_{M} ] $, we can define the softmax cross-entropy loss and the triplet for the source labeled training dataset as described in Eq. \ref{triplet} and \ref{cls}.
Then, we train  $E$ by optimizing the total loss function on source domain $L^S$ given by:

\begin{equation}
  L^S = L_{cls}^S + L_{tri}^S
\label{source_loss}
\end{equation}

\subsection{Source-guided pseudo-labeling UDA}

\subsubsection{Noisy label regularization}

In order to further exploit the source data and their labels, we propose the learning of source identity discriminative features by the feature encoder. This auxiliary task is added to the main task of learning target identity discriminative features with pseudo-labels. Since the source labels are ground-truth, this source-based auxiliary task is expected to guide the training in order to reduce the negative impact of over-fitting on erroneous target pseudo-labels, by reducting the amount of noisy data in this total (source + target) training set compared to the commonly-used target only set.

\subsubsection{Two-branch architecture}

To avoid biasing the model and thus the discriminative power of the target re-ID features with the source data, we choose a two-branch neural network architecture for the feature encoders as illustrated on Fig~\ref{fig:framework}. It is composed of a domain-shared encoder $E^C$ for low and mid-level features and two domain-specific encoders $E^S$ and $E_{T}$ resp. for source and target high level features. Our choice of modeling is supported by work that shows that features specialize for tasks in the top layers of the network.
Our feature encoder learns two separate domain-specific feature spaces given by $E^S \circ E^C$ for the source and $E_{T} \circ E^C$ for the target.

\subsubsection{Domain-specific batch-normalization}
Common neural network architectures for re-ID contain batch normalization to improve the training convergence. Experiments from the paper \cite{zajkac2019split} shows that domain shift in data can reduce performance if the statistics of batch normalization layers are not computed separately for each domain. Since we learn with data from two different domains, we follow the paper suggestion in our pipeline and compute statistics separately for source and target data.

\subsubsection{Our source-guided optimization criterion}

Similarly to the previous section, we define for the target pseudo-labeled data, a target domain classifier $W^T$ and then analogously $L^T$ \ref{source_loss} for the target samples by considering the pseudo-labels as ground-truth labels. The optimization criterion of our Source-Guided Pseudo-labeling domain adaptation framework is given by:
\begin{equation}
    L = L^S + L^{T}
\label{loss}
\end{equation}

We can note that we do not introduce any additional hyperparameter pondering the source and target term. Besides the difficulty of hyperparameter estimation in the UDA setting as mentioned earlier, our framework gives positive experimental results as it is on various datasets \ref{exp}.
The complete optimization procedure of our framework is detailed in the general Algorithm \ref{algo}, where we highlight in \textbf{bold} how our source-guided baseline differentiates from the classical target-only pseudo-labeling UDA.

\begin{algorithm}
\caption{Source-guided Pseudo-label Domain Adaptation}
\begin{algorithmic}[1]
\REQUIRE{Labeled source data $D^S = (X^S,Y^S)$, unlabeled target data $D^T = X^T$, clustering algorithm C, a source re-ID loss function $L^S$, a target re-ID loss function $L^T$, number of training epochs $N_{epoch}$, number of pseudo-labeling iterations $N_{iter}$, an initial encoder $E^{(0)}$, a \textbf{two-branch} encoder $E$}
\STATE Train the initial encoder $E^{(0)}$ on $D^S$ by optimizing $L^S(E(X^S),Y^S)$
\STATE Initialize \textbf{two-branch} $E$ such that $\mathbf{E^S \circ E^C = E^{(0)}}$ and $E^T \circ E^C = E^{(0)}$
\FOR{$t=1$ to $N_{iter}$}
\STATE Compute target features: $F^T  \leftarrow E(X^T)$
\STATE Compute pairwise target feature distances: $d(F^T) \leftarrow d(F^T_i,F^T_j)_{1 \leq i,j \leq N_{T}}$ 
\STATE Pseudo-label some/all target samples by clustering: $ (X^T,\hat{Y^T}) \leftarrow C(d(F^T),D^T)$
\STATE Train E during $N_{epoch}$ by optimizing $\mathbf{L^S((E^S \circ E^C)(X^S), Y^S) +}  L^{T}((E^T \circ E^C)(X^T), \hat{Y}^T)$
\ENDFOR
\STATE Return $E^T \circ E^C$
\end{algorithmic}
\label{algo}

\end{algorithm}

\section{Experiments}
\label{exp}

\subsection{Datasets}

We evaluate our framework on three commonly-used re-ID datasets: Market-1501 (Market) \cite{zheng2015scalable}, DukeMTMC-re-ID (Duke) \cite{ristani2016performance} and MSMT17 (MSMT) \cite{wei2018person}. Market-1501 and DukeMTMC-re-ID are the most commonly-used datasets for re-ID evaluation. Each of them defines a domain.\\
\indent Market-1501 is composed of 32,668 labeled images from 1501 people captured by 6 outdoor cameras. It is divided into a training set of 12,936 images of 751 identities and a test set with 19,732 images of 750 identities different from the training ones.\\
\indent DukeMTMC-re-ID contains 36,411 labeled images of 702 IDs taken by 8 outdoor cameras. It is split into a training set with 6,522 images of 702 identities and 19,889 images of 702 other identities for the test set.\\
\indent MSMT17 is a larger dataset, with 126,441 labeled images of 4,101 identities collected by 15 indoor and outdoor cameras. The training set contains 32,621 images of 1,041 identities and the testing set 93,820 images of 3,060 other identities.
It is worth noticing that MSMT17 is a much more challenging dataset than the other two: due to the size of its test set, its number of identities and cameras, MSMT17 is the closest dataset to the conditions of a large-scale re-ID system deployment. Therefore, re-ID UDA from Market or Duke to a larger-scale dataset such as MSMT is more difficult but also more interesting since in practice we often have few labeled data and a lot of unlaballed data.\\
\indent We evaluate our framework on the commonly tested Duke-to-Market (Duke being the labeled source and Market the unlabeled target dataset) and Market-to-Duke UDA tasks. Besides, we test our approach on the more challenging adaptation tasks Market-to-MSMT and Duke-to-MSMT UDA tasks. Mean average precision (mAP) and CMC top-1 accuracy are reported to measure our framework's performance.

\subsection{Experimental settings}

\begin{figure}[t]
\centering
\includegraphics[width=\columnwidth]{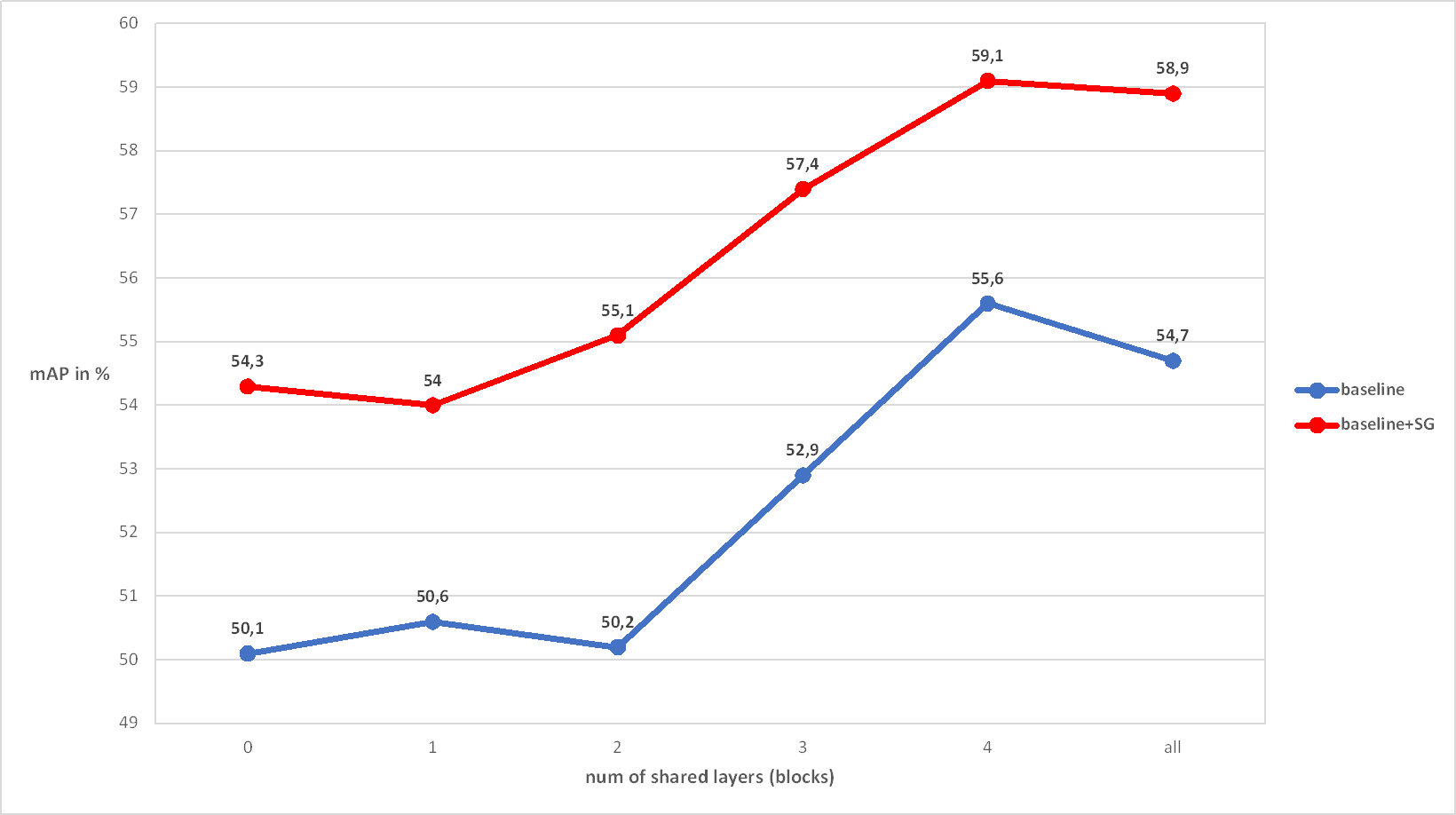}
\caption{Impact on mAP (in $\%$) of the number of shared layers used in the shared encoder $E^C$ of baseline+SG, on Duke-to-Market and Market-to-Duke.}
\label{layers}
\end{figure}

\subsubsection{Tested models}

To show that our guideline can easily be added and contribute to various pseudo-labeling UDA approaches, we choose to integrate it into two target-only frameworks:
\begin{itemize}
    \item Our baseline: The classical pseudo-labeling UDA algorithm based on the UDAP approach \cite{song2020unsupervised} which is not designed to be robust to overfiting pseudo-labels'errors. It corresponds to the MMT \cite{ge2020mutual} framework without Mutual Mean Teaching and k-means replaced by DBSCAN clustering algorithm on the pairwise matrix of k-reciprocal encoding distances \cite{zhong2017re} of the target training features.
    \item Our baseline+MMT \cite{ge2020mutual} framework: actually the best state-of-the-art  pseudo-labeling UDA method which mitigates for bad effects due to pseudo-label errors. It uses k-means as the clusterer in the feature space to predict target pseudo-labels based on the parirwise L2 distance matrix of the target training set features, as well as the mutual-mean teaching strategy described in their paper for pseudo-label error robustness.
\end{itemize}

\subsubsection{Implementation details}
\label{impl}

\textbf{Initialization phase.} We follow the guidelines for supervised training from the paper \cite{luo2019bag} adopted by MMT \cite{ge2020mutual}.For fair comparison, we choose the ResNet-50 \cite{he2016deep} initialized on the pretrained ImageNet weights \cite{deng2009imagenet} as our feature extractor. We use batch of 64 images composed of 16 identities and 4 shots per identity. Images are randomly flipped and resized to 256x128. Random Erasing Data augmentation \cite{zhong2017random} is not used during the initialization phase since it may reduce transferability of the source model features to the target domain thus generating more errors in pseudo-labels after UDA initialization \cite{luo2019bag}. We use ADAM as the optimizer and a weight decay of $5 \cdot 10^{-4}$. The initial learning rate is set to 0.00035 and is decreased to 1/10 of its previous value on the 40th and 70th epoch in a total of 80 epochs.\\
\indent \textbf{Pseudo-labeling phase.} Unless otherwise specified, we use as a common encoder all but the layers from the last convolutional block and after (4 first blocs of layers), as motivated by our parameter analysis \ref{layers}. We use the same initialization phase preprocessing with two batches of 64 images, adding Random Erasing Data augmentation \cite{zhong2017random}: one for source images and another one for target. We feed separately the network with the source and target batches to ensure domain-specific batch normalization statistics as explained in Section \ref{sec:framework}. For our baseline, after each uour baselinete phase of the pseudo-labels, the pseudo-ID and their number can change. Therefore we initialize randomly a new classification layer for target after each pseudo-labels our baseline.
Other hyperparameters (clusterer parameter, triplet loss margin, number of iterations for pseudo-labeling,...) used after the initialization phase are kept the same as the UDA paper's ones (resp MMT's ones): they correspond to the best hyperparameters found after validation on the target test set in their papers.\\
\indent \textbf{Source-Guided (SG) versions.} For the baseline+SG framework, it is the direct application of our source-guided Algorithm \ref{algo}. 
For our baseline+MMT+SG, the two-branch architecture is adopted for the two models (mean teacher and student) that train mutually in their baseline. It implies that we do not modify the soft and hard label loss weighting and we directly add the source term (on hard labels) as described in Eq. \ref{loss}. 

\section{Results}
\label{Results}

In this part, we conduct parameter analysis and comparison with existing state-of-the-art re-ID UDA methods.

    \begin{figure}[t!]
    \centering
    \includegraphics[width=\columnwidth]{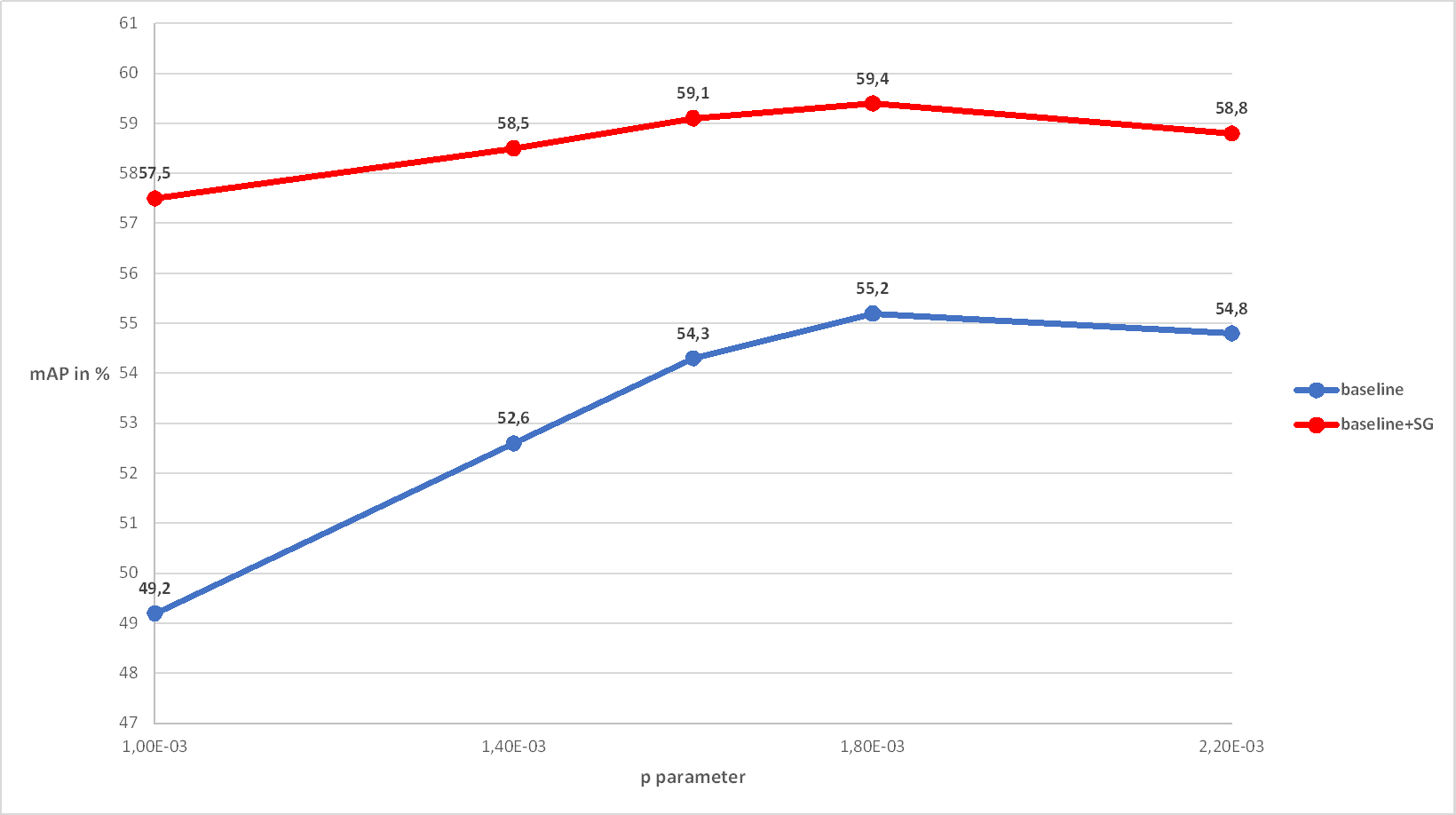}
    \caption{Robustness of our baseline+SG to $p$ parameter's changes ($p$ controls DBSCAN neighborhood distance parameter) compared to the target-only framework our baseline on Duke-to-Market.}
    \label{p_d2m}
    \end{figure}
    
    \begin{figure}[t]
    \centering
    \includegraphics[width=\columnwidth]{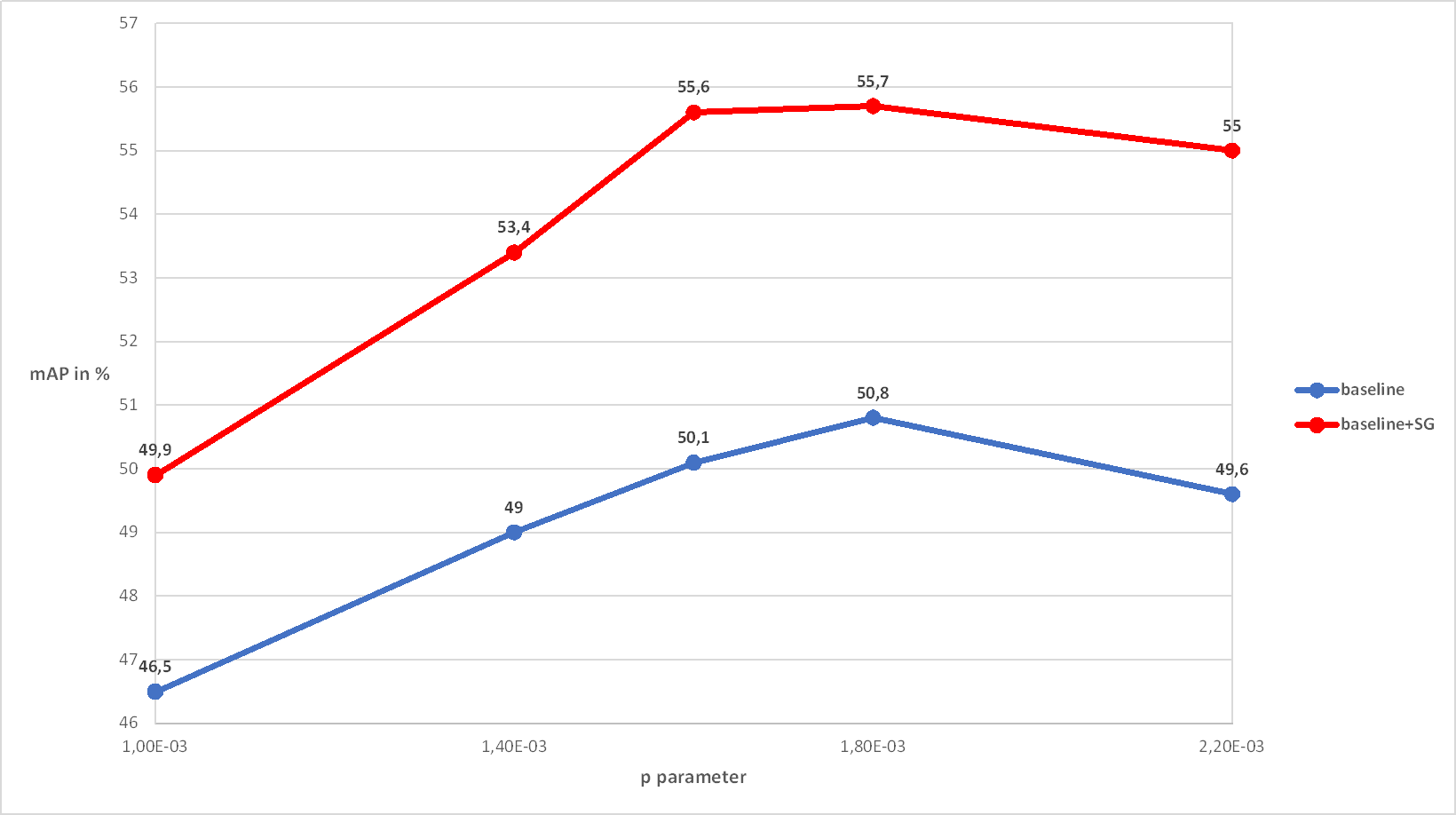}
    \caption{Robustness of our baseline+SG to $p$ parameter's changes ($p$ controls DBSCAN neighborhood distance parameter) compared to the target-only framework our baseline on Market-to-Duke.}
    \label{p_m2d}
    \end{figure}
    
    \begin{figure}[t!]
    \centering
    \includegraphics[width=\columnwidth]{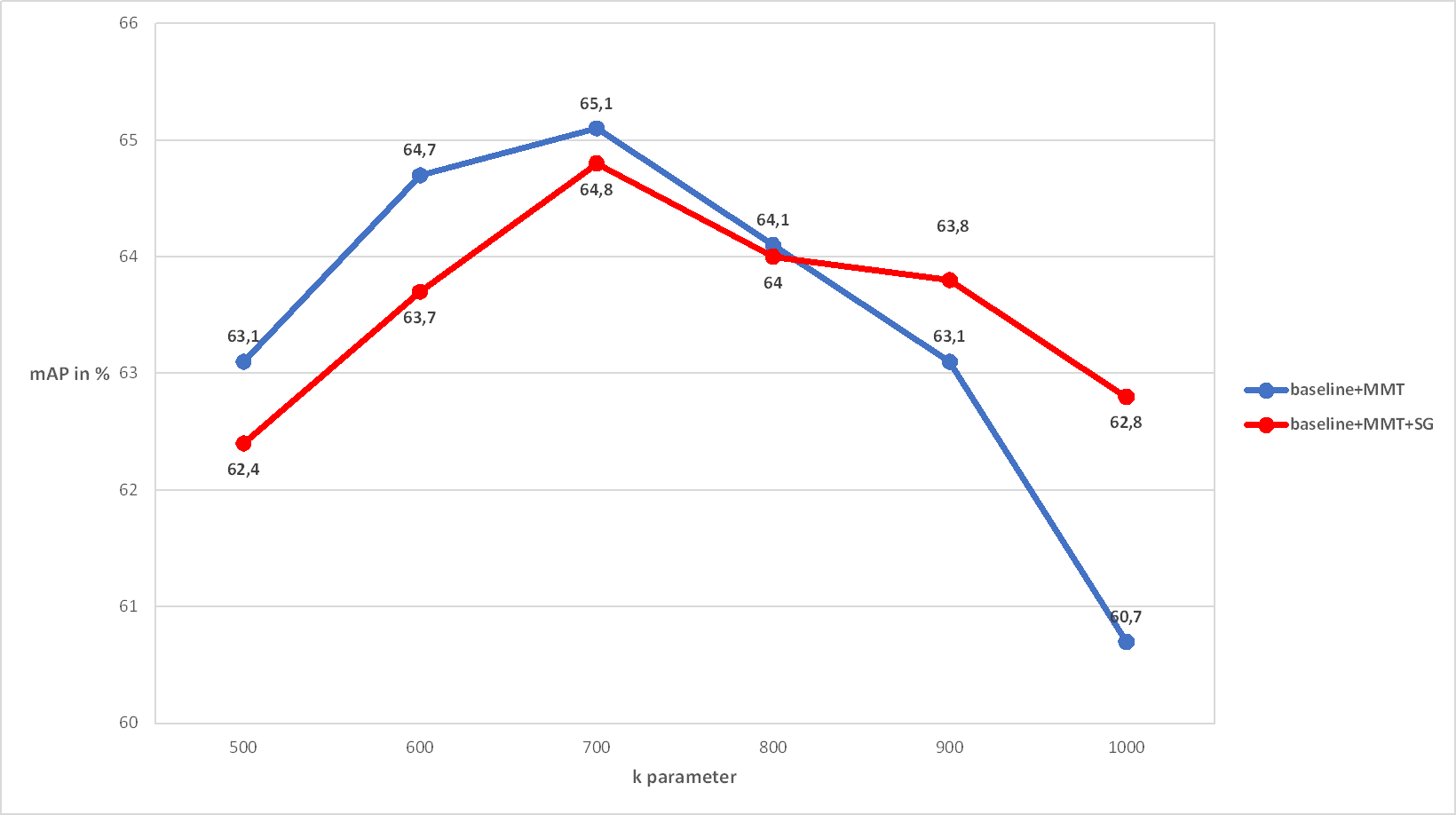}
    \caption{Robustness of our baseline+MMT+SG to $k$ parameter's changes ($k$ controls k-means number of clusters) compared to the target-only framework MMT on Market-to-Duke.}
    \label{mmt_m2d}
    \end{figure}
    
    \begin{figure}[t!]
    \centering
    \includegraphics[width=\columnwidth]{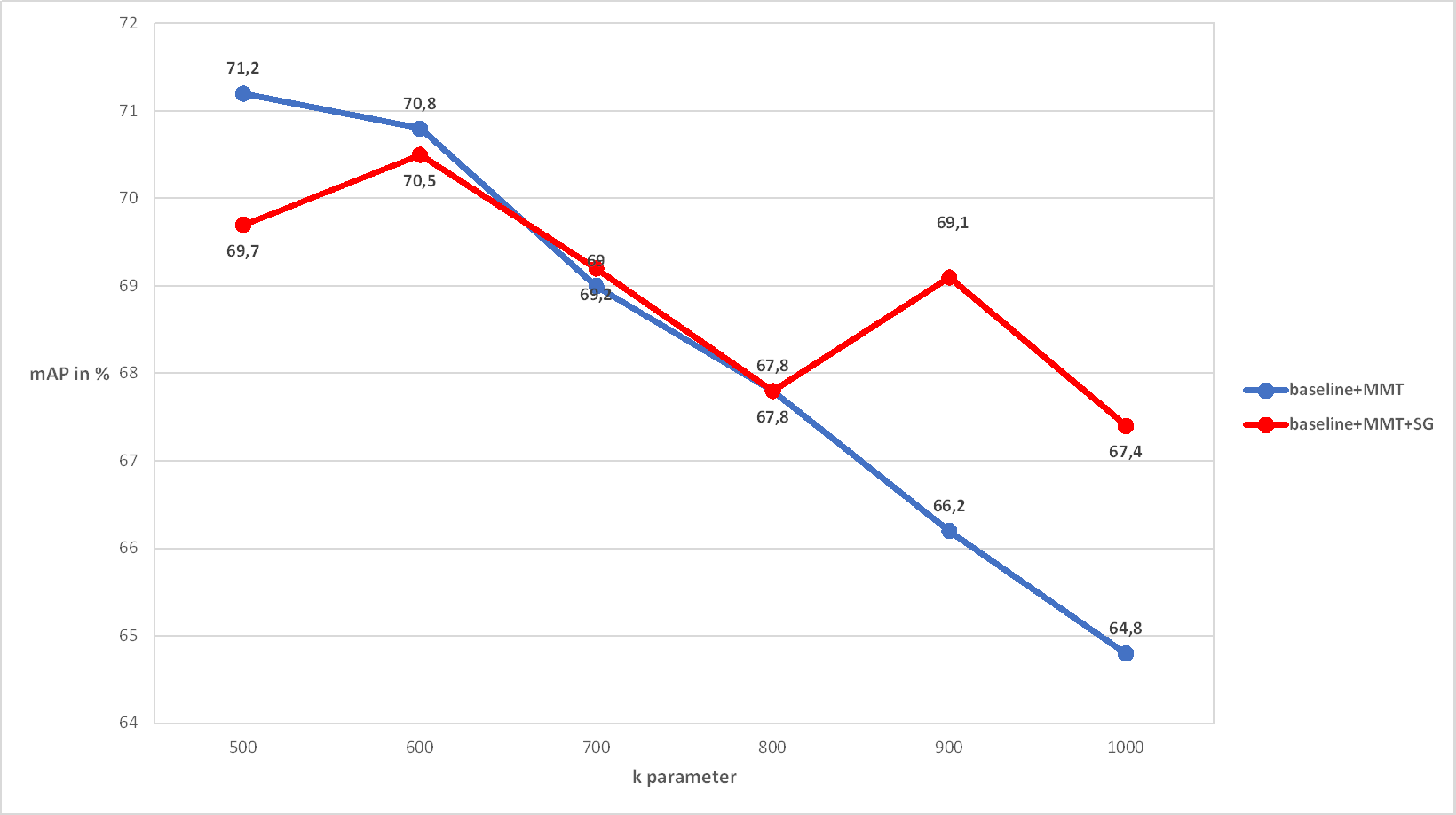}
    \caption{Robustness of our our baseline+MMT+SG to $k$ parameter's changes ($k$ controls k-means number of clusters) compared to the target-only framework MMT on Duke-to-Market.}
    \label{mmt_d2m}
    \end{figure}
    
    \begin{figure}[t!]
    \centering
    \includegraphics[width=\columnwidth]{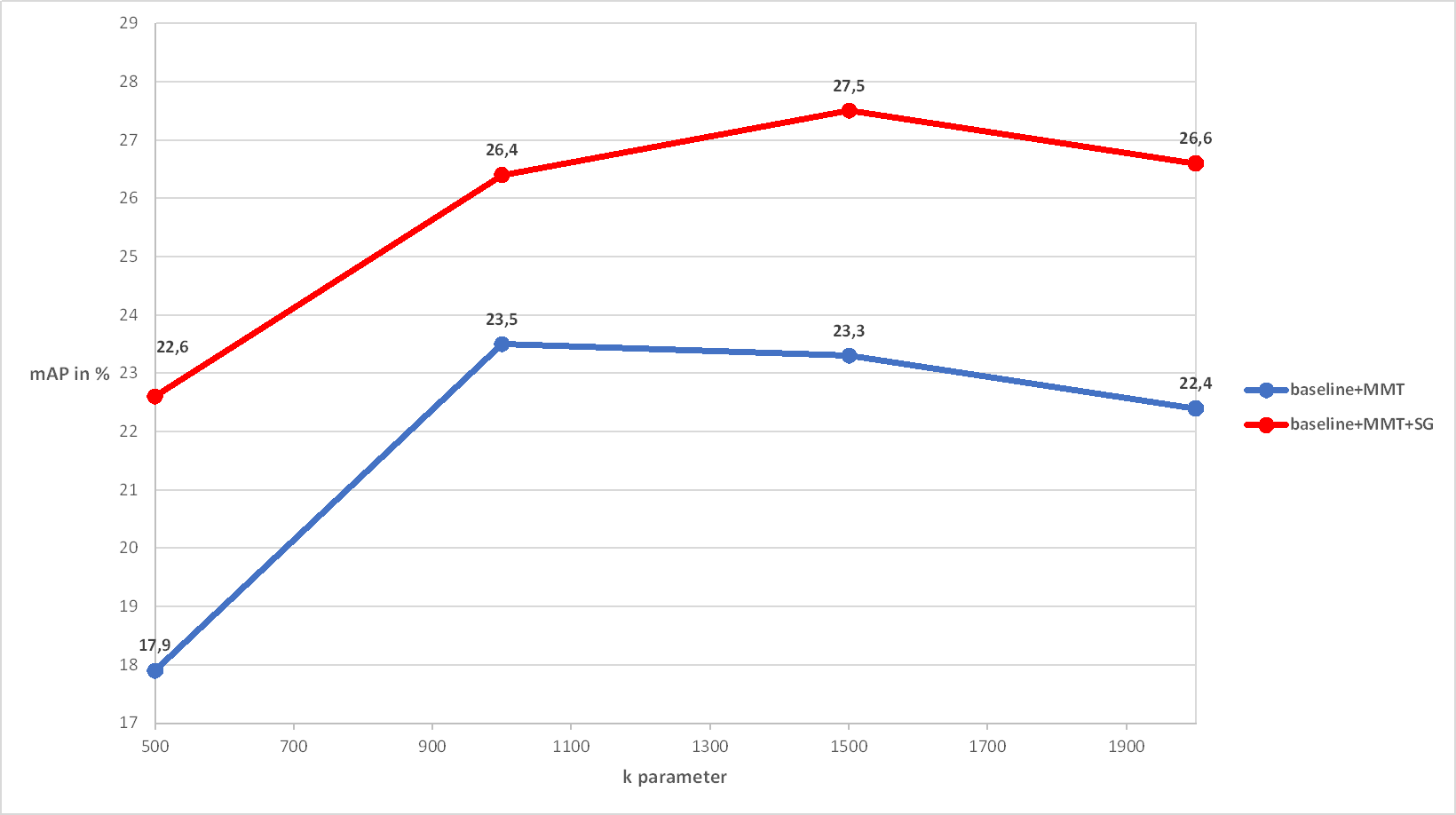}
    \caption{Robustness of our baseline+MMT+SG to $k$ parameter's changes ($k$ controls k-means number of clusters) compared to the target-only framework MMT on Duke-to-MSMT.}
    \label{p_d2msmt}
    \end{figure}
    
    \begin{figure}[t!]
    \centering
    \includegraphics[width=\columnwidth]{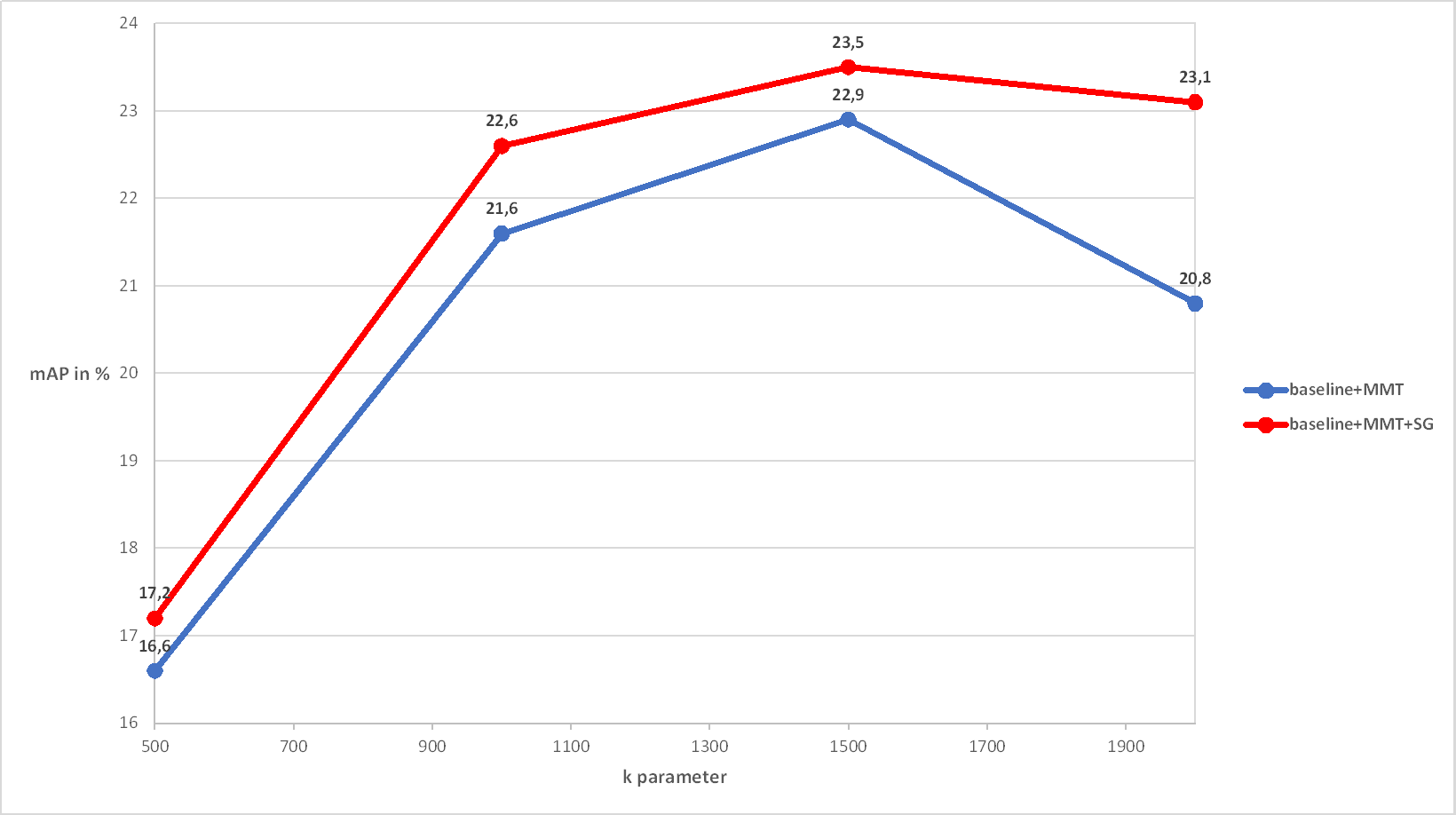}
    \caption{Robustness of our baseline+MMT+SG to $k$ parameter's changes ($k$ controls k-means number of clusters) compared to the target-only framework MMT on Market-to-MSMT.}
    \label{p_m2msmt}
    \end{figure}

\subsection{Parameter analysis.}

\subsubsection{Do the source domain help the model to learn better target features ?}
As explained and motivated in Section \ref{sec:framework}, we propose a two-branch architecture to learn domain-specific high level ID discriminative features based on low and mid level domain-shared features learned with labeled source data and pseudo-labeled target data. We can wonder if our two-branch encoder manages to leverage the source samples to improve the target features. Furthermore, we would like to know how many layers we should share to take advantage from the labeled source data without negatively biasing the target features.\\
To answer these two questions, we vary the number of ResNet-50 layers shared between source and target domain through the $E^C$ encoder of our Source-Guided baseline. The ResNet-50 architecture can be divided into $5$ convolutional blocks of layers defined in the ResNet paper \cite{he2016deep} to which we refer to vary the number of shared layers. Case "$0$ shared layer" corresponds to the classical target-only pseudo-labeling methods (Figure \ref{fig:framework}(1)) which corresponds to our baseline, where case "$5$ shared layers" to sharing the whole ResNet50 between source and target domain. 
\indent Experiments show on Fig.\ref{layers} increasing performances when the number shared layers increase. More precisely, the best mAP is reached for $4$ shared block of layers: $59.1\%$ mAP for Duke-to-Market and $55.6\%$ mAP for Market-to-Duke, increasing resp. the performances by $4.8 p.p.$ and $5.5 p.p.$ compared to the target only model. Our source-guided baseline outperforms the classical target-only pseudo-labeling baseline and our partially-shared strategy gives the best results for Duke-to-Market and Market-to-Duke.
Moreover, we also notice that the slight drop of performances for a fully-shared model may highlight a trade-off between:
\begin{itemize}
    \item Sharing the highest level (specifically the last one) layers which can benefit the most from the auxiliary source re-ID task regularization
    \item Biasing the highest level feature space which is directly used for re-ID on target domain.
\end{itemize}

\subsubsection{Efficiency of Specific Batch Normalization}

\begin{table}[t]
\caption{Impact of domain-specific batch-normalization on domain adaptation performance (mAP in $\%$) .}
\label{batchnorm}
\begin{tabular}{|c|c|c|}
\hline
\multirow{2}{*}{Methods}        & Market-to-Duke & Duke-to-Market \\ \cline{2-3} 
                                & mAP            & mAP            \\ \hline
baseline                             & 50.1           & 54.3           \\
Domain-Shared BatchNorm baseline+SG   & 36.7           & 43.1           \\
Domain-Specific BatchNorm baseline+SG & \textbf{55.6} & \textbf{59.1} \\ \hline
\end{tabular}
\end{table}

We study the effectiveness of domain-specific batch nornmalization as motivated in Section \ref{sec:framework}. \. To do so, we compare our baseline+SG framework to a version that shares the batch normalization between domains (Domain Shared BatchNorm baseline+SG). From Table \ref{batchnorm}, we notice that sharing the batch normalization deteriorates the performance on both couples of Market-to-Duke and Duke-to-Market adaptation datasets. mAP drop more than $-10$ p.p. below the model using only the target data (our baseline). Only the addition of domain-specific batch normalization increases the performances of our framework above the our baseline model. These experiments therefore show that the use of domain-specific batch normalization is an essential key of our framework in order not to deteriorate the learning of discriminative target features by biasing the batch normalization statistics.

\subsubsection{Is our strategy of using source samples robust to clustering parameters changes ?}

\indent In the UDA setting, choosing or tuning hyperparameter is a tricky task due to the absence of a labeled validation set for the target domain. It is therefore important in practice to design UDA methods robust to hyperparameter changes. In particular, pseudo-labeling UDA methods \cite{song2020unsupervised} \cite{ge2020mutual} give experimental evidences that performance can be very sensible to clustering parameters changes. That's why we would like to focus in this part on the performance of our source-guided frameworks when these clustering parameters change. For the baseline+SG baseline, we study the $p$ parameter of DBSCAN as defined and studied in UDAP paper \cite{song2020unsupervised} and for our baseline+MMT+SG we focus on the $k$ paremeter of k-means as in the MMT paper \cite{ge2020mutual}. We compare our two frameworks to their target-only versions.
\indent The $p$ parameter directly controls the DBSCAN neighborhood distance parameter in our baseline and baseline+SG: it determines the percentage of the smallest distances to be used to define clusters in the feature space.
We vary $p$ in the same interval of the UDAP paper \cite{song2020unsupervised}. By doing so, we notice that for Duke-to-Market in Figure \ref{p_d2m} and Market-to-Duke in Figure  \ref{p_m2d}, in baseline+SG the source benefits the domain adaptation by pseudo-labeling: we go from $50.6\%$ to $55.9\%$ on Market-to-Duke and from $54.6\%$ to $59.4\%$ by comparison to the target only version. For every parameter value $p$, there is an improvement in performance for baseline+SG ranging from at least $3.4$ p.p. for Market-to-Duke and $4$ p.p. for Duke-to-Market in comparison to the target only baseline our baseline: *** our framework seems to be robust to the change of parameters $p$.
\indent The $k$ parameter determines the number of clusters in the MMT frameworks. We choose the same interval of values as in the MMT paper \cite{ge2020mutual} for varying the $k$ parameter.
For Maket-to-Duke and Duke-to-Market in Figures \ref{mmt_m2d} and \ref{mmt_d2m}, the addition of the source term with our baseline+MMT+SG does not seem to increase the maximum performance, which is reached by the classical target-only MMT model. Nevertheless, there are quite different performance curve trends between MMT and our baseline+MMT+SG. our baseline+MMT+SG seems to be more robust for $k$ values above 800 (Market and Duke containing 751 and 702 identities respectively), i.e. when a number of clusters is chosen above the actual number of identities. While MMT already proposes a strategy of resistance to pseudo-label noise, which can explain the non improvement of the best mAP, the addition of the source-guidance in our baseline+MMT+SG seems to confer more stability to the clusterer parameter variations. Numerically, this stability can be observed by calculating the standard deviation (std) of the mAPs over $k$: using the source, we go from $2.5$ p.p. to $1.2$ p.p. on Duke-to-Market and from $1.6$ p.p. to $0.9$ p.p. on Market-to-Duke, with on average equivalent performance for the two pairs of data sets. This stability conferred by the source is interesting given that we not know the number of identities of the training set target, which can only be estimated at best. We believe that the number of ground-truth ID for the source dataset might help to reduce the impact on training of the cluster surplus of the target set.\\
\indent In the more challenging cases where MSMT is the target dataset, there is a clear contribution from the source. We can see in Figures \ref{p_m2msmt} and \ref{p_d2msmt} that it is stable to the change in $k$ and allowed to increase the maximum performance: from $23.5$\% to $27.5$\% for Duke-to-MSMT and from $22.9$\% to $23.5$\% for Market-to-MSMT. There is also a higher source contribution at high $k$ values. It can be assumed that our baseline+MMT+SG works better in this more challenging case of adaptation because of the presence of more noisy labels during the transfer of the source model for initialization of pseudo-labeling: adding our strategy of exploiting source data therefore presents less redundancy with the one already implemented in the MMT framework, and even more if we "over-estimate" the number of clusters.

\subsection{Comparison with state-of-the-art methods}
\label{sec:sota}

\begin{table}[t]
\begin{center}
\caption{Comparison with state-of-the-art methods.}
\label{table^Sota}
\begin{tabular}{ccccc}
\hline
\multicolumn{1}{|c|}{\multirow{2}{*}{Methods}} & \multicolumn{2}{c|}{Market-to-Duke}                & \multicolumn{2}{c|}{Duke-to-Market}                \\ \cline{2-5} 
\multicolumn{1}{|c|}{}           & mAP                  & \multicolumn{1}{c|}{top-1} & mAP                  & \multicolumn{1}{c|}{top-1} \\ \hline
\multicolumn{1}{|c|}{SPGAN \cite{deng2018image}}      & 22.3                 & \multicolumn{1}{c|}{41.1}  & 22.8                 & \multicolumn{1}{c|}{51.5}  \\
\multicolumn{1}{|c|}{TJ-AIDL \cite{wang2018transferable}}    & 23.0                 & \multicolumn{1}{c|}{44.3}  & 26.5                 & \multicolumn{1}{c|}{58.2}  \\
\multicolumn{1}{|c|}{MMFA \cite{lin2018multi}}       & 24.7                 & \multicolumn{1}{c|}{45.3}  & 38.3                 & \multicolumn{1}{c|}{66.2}  \\
\multicolumn{1}{|c|}{HHL \cite{zhong2018generalizing}}        & 27.2                 & \multicolumn{1}{c|}{46.9}  & 31.4                 & \multicolumn{1}{c|}{62.2}  \\
\multicolumn{1}{|c|}{CFSM \cite{chang2019disjoint}}       & 27.3                 & \multicolumn{1}{c|}{49.8}  & 28.3                 & \multicolumn{1}{c|}{61.2}  \\
\multicolumn{1}{|c|}{UCDA-CCE \cite{qi2019novel}}   & 31.0                 & \multicolumn{1}{c|}{47.7}  & 30.9                 & \multicolumn{1}{c|}{60.4}  \\
\multicolumn{1}{|c|}{ARN \cite{li2018adaptation}}        & 33.4                 & \multicolumn{1}{c|}{60.2}  & 39.4                 & \multicolumn{1}{c|}{70.3}  \\
\multicolumn{1}{|c|}{ECN \cite{zhong19enc}}        & 40.4                 & \multicolumn{1}{c|}{63.3}  & 43.0                 & \multicolumn{1}{c|}{75.1}  \\
\multicolumn{1}{|c|}{PoseDA-Net \cite{li2019cross}}    & 45.1                 & \multicolumn{1}{c|}{63.2}  & 47.6                 & \multicolumn{1}{c|}{75.2}  \\
\multicolumn{1}{|c|}{UDAP \cite{song2020unsupervised}}       & 49.0                 & \multicolumn{1}{c|}{68.4}  & 53.7                 & \multicolumn{1}{c|}{75.8}  \\
\multicolumn{1}{|c|}{SSG \cite{fu2019self}}        & 53.4                 & \multicolumn{1}{c|}{73.0}  & 58.3                 & \multicolumn{1}{c|}{80.0}  \\
\multicolumn{1}{|c|}{ISSDA-re-ID \cite{tang2019unsupervised}} & 54.1                 & \multicolumn{1}{c|}{72.8}  & 63.1                 & \multicolumn{1}{c|}{81.3}  \\
\multicolumn{1}{|c|}{PCB-PAST \cite{zhang2019self}}   & 54.3                 & \multicolumn{1}{c|}{72.4}  & 54.6                 & \multicolumn{1}{c|}{78.4}  \\
\multicolumn{1}{|c|}{ACT \cite{yang2019asymmetric}}        & 54.5                 & \multicolumn{1}{c|}{72.4}  & 60.6                 & \multicolumn{1}{c|}{80.5}  \\
\multicolumn{1}{|c|}{MMT \cite{ge2020mutual}}        & \textbf{65.1}        & \multicolumn{1}{c|}{78.0}  & \textbf{71.2}        & \multicolumn{1}{c|}{87.7}  \\ 
\multicolumn{1}{|c|}{Our (target-only) baseline}  & 50.1                 & \multicolumn{1}{c|}{70.1}  & 54.3                 & \multicolumn{1}{c|}{73.5}  \\ \hline
\multicolumn{1}{|c|}{Our baseline+SG}  & 55.6                 & \multicolumn{1}{c|}{73.2}  & 59.1                 & \multicolumn{1}{c|}{80.8}  \\
\multicolumn{1}{|c|}{Our baseline+MMT+SG}                & 64.8          & \multicolumn{1}{c|}{\textbf{78.5}} & 70.5          & \multicolumn{1}{c|}{\textbf{88.1}} \\ \hline
\multicolumn{1}{l}{}             & \multicolumn{1}{l}{} & \multicolumn{1}{l}{}       & \multicolumn{1}{l}{} & \multicolumn{1}{l}{}       \\ \hline
\multicolumn{1}{|c|}{\multirow{2}{*}{Methods}} & \multicolumn{2}{c|}{Market-to-MSMT}                & \multicolumn{2}{c|}{Duke-to-MSMT}                  \\ \cline{2-5} 
\multicolumn{1}{|c|}{}           & mAP                  & \multicolumn{1}{c|}{top-1} & mAP                  & \multicolumn{1}{c|}{top-1} \\ \hline
\multicolumn{1}{|c|}{PTGAN \cite{wei2018person}}      & 2.9                  & \multicolumn{1}{c|}{10.2}  & 3.3                  & \multicolumn{1}{c|}{11.8}  \\
\multicolumn{1}{|c|}{ECN \cite{qi2019novel}}        & 8.5                  & \multicolumn{1}{c|}{25.3}  & 10.2                 & \multicolumn{1}{c|}{30.2}  \\
\multicolumn{1}{|c|}{UDAP \cite{song2020unsupervised}}       & 12.0                 & \multicolumn{1}{c|}{30.5}  & 16.0                 & \multicolumn{1}{c|}{39.2}  \\
\multicolumn{1}{|c|}{SSG \cite{fu2019self}}        & 13.2                 & \multicolumn{1}{c|}{49.6}  & 13.3                 & \multicolumn{1}{c|}{32.2}  \\
\multicolumn{1}{|c|}{MMT \cite{ge2020mutual}}        & 22.9                 & \multicolumn{1}{c|}{49.2}  & 23.5                 & \multicolumn{1}{c|}{50.1}  \\ 
\multicolumn{1}{|c|}{Our (target-only) baseline}  & 11.6                 & \multicolumn{1}{c|}{29.8}  & 14.8                 & \multicolumn{1}{c|}{36.1}  \\ \hline
\multicolumn{1}{|c|}{Our baseline+SG}  & 14.9                 & \multicolumn{1}{c|}{35.4}  & 19.3                 & \multicolumn{1}{c|}{45.6}  \\
\multicolumn{1}{|c|}{Our baseline+MMT+SG}                & \textbf{23.5} & \multicolumn{1}{c|}{\textbf{50.2}} & \textbf{27.5} & \multicolumn{1}{c|}{\textbf{56.1}} \\ \hline
\end{tabular}
\end{center}
\end{table}

\indent We compare in Table~\ref{table^Sota} our two source-guided frameworks with the state of the art on Duke-to-Market, Market-to-Duke, Market-to-MSMT and Duke-to-MSMT.\\
\indent On Market-to-Duke and Duke-to-Market, our two frameworks far exceed those that do not use pseudo-labels at all. Our frameworks outperform all state of the art methods except MMT on Market-to-Duke and Duke-to-Market. Specifically, baseline+SG outperforms similar Pseudo-labeling methods that do not exploit the labeled source data after pseudo-label initialization: UDAP and PCB-PAST, but also those that integrate noise resistance or error filtering strategies in pseudo-labels such as ISSDA-re-ID, SSG and ACT on Maket-to-Duke. On Duke-to-Market, ISSDA-re-ID surpasses baseline+SG, even if baseline+SG has the advantage of simple integration in any framework and needs few additional parameters (no need to train a CycleGAN or a second model for asymmetric co-teaching).\\
Even if our baseline+MMT+SG, has a slightly lower maximum mAP than MMT, it offers the advantage of a better stability on the parameter of the number of clusters (estimated) as seen in the parameter analysis part, which is a major asset in domain adaptation.\\
\indent On Market-to-MSMT and Duke-to-MSMT, our baseline+SG framework exceeds all methods except MMT. In particular, it exceeds SSG by $+1.7$ p.p. mAP on Market-to-MSMT and UDAP by $+3.3$ p.p. on Duke-to-MSMT. The our baseline+MMT+SG framework exceeds the state of the art on these two pairs of data sets: on Market-to-MSMT and Duke-to-MSMT, our baseline+MMT+SG resp. increases performance by $+0.6$ p.p. and by $+4$ p.p. on Duke-to-MSMT.

\section{Conclusion}
In this paper, we propose a guideline to leverage the commonly under-used ground-truth labeled source samples during pseudo-labeling domain adaptation. It consists of guiding the target pseudo-label training stage with an auxiliary ID-discriminative source feature learning task, while preventing the source samples from biasing the training by using domain-specific batch normalization and an architecture with  two domain-specific branches. Experiments on combining our framework with two state-of-the-art methods are carried out on different datasets. They show that leveraging the source samples brings more stability with relation to the choice of clustering parameters and improves performance on particularly challenging adaptation settings.

\printbibliography

\end{document}